\documentclass[journal,transmag]{IEEEtran}

\ifCLASSINFOpdf
\else
\fi
\usepackage{graphicx}
\usepackage{amsmath}
\usepackage{cuted}
\usepackage{algorithm}
\usepackage{algorithmicx}
\usepackage{algpseudocode}

\usepackage{multicol}
\usepackage{multirow}
\usepackage{hyperref}
\usepackage{pifont}
\usepackage{siunitx}
\usepackage{array}
\usepackage{tabularray} 
\usepackage{makecell} 
\usepackage{threeparttable}
\usepackage{xcolor}
\usepackage{gensymb}
\begin{document}
%
\title{SurfAAV: Design and Implementation of a Novel Multimodal Surfing Aquatic-Aerial Vehicle}


\author{\IEEEauthorblockN{Kun Liu\IEEEauthorrefmark{1},
Junhao Xiao\IEEEauthorrefmark{1},
Hao Lin\IEEEauthorrefmark{1},
Yue Cao\IEEEauthorrefmark{1},
Hui Peng\IEEEauthorrefmark{2},
Kaihong Huang\IEEEauthorrefmark{1*}, and
Huimin Lu\IEEEauthorrefmark{1}}
\IEEEauthorblockA{\IEEEauthorrefmark{1}College of Intelligence Science and Technology, National University of Defense Technology, Changsha, China}
\IEEEauthorblockA{\IEEEauthorrefmark{2}School of Electronic Information, Central South University, Changsha, China}
\thanks{Manuscript received xx xx, 2012; revised xx xx, 2015. 
Corresponding author: xx (email: xxx).}}

\markboth{Journal of \LaTeX\ Class Files,~Vol.~14, No.~8, August~2015}%
{Shell \MakeLowercase{\textit{et al.}}: Bare Demo}
%



\IEEEtitleabstractindextext{%
\begin{abstract}
Despite significant advancements in the research of aquatic-aerial robots, existing configurations struggle to efficiently perform underwater, surface, and aerial movement simultaneously. In this paper, we propose a novel multimodal surfing aquatic-aerial vehicle, SurfAAV, which efficiently integrates underwater navigation, surface gliding, and aerial flying capabilities. Thanks to the design of the novel differential thrust vectoring hydrofoil, SurfAAV can achieve efficient surface gliding and underwater navigation without the need for a buoyancy adjustment system. This design provides flexible operational capabilities for both surface and underwater tasks, enabling the robot to quickly carry out underwater monitoring activities. Additionally, when it is necessary to reach another water body, SurfAAV can switch to aerial mode through a gliding takeoff, flying to the target water area to perform corresponding tasks. The main contribution of this letter lies in proposing a new solution for underwater, surface, and aerial movement, designing a novel hybrid prototype concept, developing the required control laws, and validating the robot's ability to successfully perform surface gliding and gliding takeoff. SurfAAV achieves a maximum surface gliding speed of 7.96 m/s and a maximum underwater speed of 3.1 m/s. The prototype's surface gliding maneuverability and underwater cruising maneuverability both exceed those of existing aquatic-aerial vehicles.
\end{abstract}

}

\maketitle

\IEEEdisplaynontitleabstractindextext

%
\IEEEpeerreviewmaketitle

\section{Introduction}

\IEEEPARstart{I}{n}
recent years, with the rapid development of robotics technology, unmanned aquatic-aerial vehicles(UAAVs) capable of adapting to complex environments and performing diversified tasks have gradually become a research hotspot. 
These robots integrate the advantages of both autonomous underwater vehicles(AUVs) and unmanned aerial vehicles(UAVs), allowing them to freely switch between motion modes in water and air. This capability greatly broadens the application scope of traditional robots, demonstrating enormous potential in multi-domain missions such as environmental monitoring\cite{ore2015autonomous}, disaster rescue\cite{murphy2008cooperative}, and national defense\cite{manyam2017multi}. 
While multi-platform collaboration can perform multi-domain tasks such as water body monitoring, it also adds complexity of mission and reduces operational reliability\cite{yao2023review}.
In contrast, UAAVs can overcome these challenges by breaking through the water surface. They can fly quickly to target areas and then conduct fine underwater inspections, thus enabling comprehensive three-dimensional monitoring of aquatic environments.

The existing UAAVs are classified mainly into four types based on their configurations: bionic UAAVs\cite{siddall2014launching, 7725996, zufferey2019consecutive, 7353394, chen2017biologically}, multi-rotor UAAVs\cite{8202261, alzu2018loon, 9196687, 10160899, 10814088}, fixed-wing UAAVs\cite{7762760, 9120363, 8039429, rockenbauer2021dipper, sun2024design}, and hybrid UAAVs\cite{lyu2022toward, jin2024nezha, 10577093, 8733846, zou2023design, qin2023aerial, wang2024design, 10844691}. 

\begin{figure}
    \centering
    \includegraphics[width=0.8\linewidth]{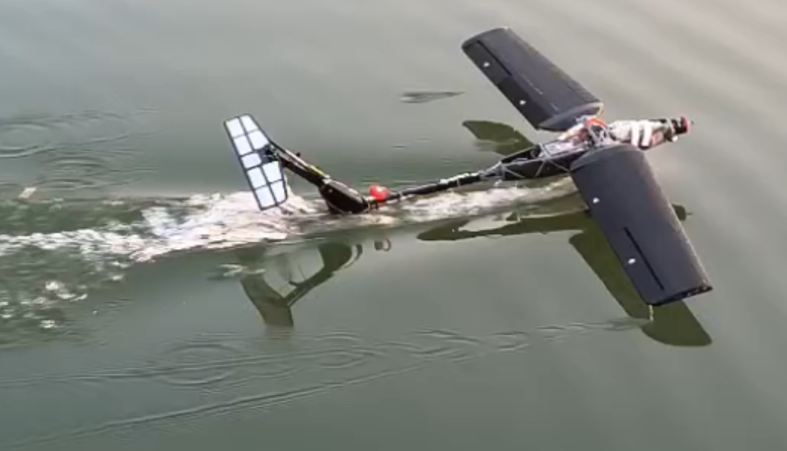}
    \caption{The SurfAAV gliding on the water surface.}
    \label{fig:water-surfing}
\end{figure}

Bionic UAAVs are primarily inspired by animals such as flying squids and birds. When a flying squid perceives danger, it jets out of the water to escape the threatening area. Inspired by this, researchers have developed flying squid-inspired UAAVs\cite{siddall2014launching, 7725996, zufferey2019consecutive} that can transition from water to air through high-pressure gas ejection. Although they utilize a fixed-wing configuration that allows for gliding in the air, they lack the capability for sustained flight.      
Additionally, researchers have noted that some bird species, such as puffins and guillemots, also demonstrate abilities to move between water and air. These birds take off from water surface by rapidly flapping their wings. Inspired by this mechanism, \cite{7353394} introduces the first insect-scale micro-robot, RoboBee, capable of both aerial flight and aquatic locomotion. However, due to the robot’s inability to overcome surface tension, its initial design could not autonomously transition from water to air. To address this limitation, \cite{chen2017biologically} presents structural improvements to the micro-robot, ultimately enabling successful takeoff from water surface. Experimental results demonstrate that the robot achieves an exit speed of 1.8 m/s during the water-exit phase.

Inspired by the vertical take-off and landing(VTOL) capabilities of multirotors, scholars have developed multi-rotor UAAVs.
These vehicles have a simple structure and are easy to control. However, they face significant resistance and reduced efficiency during underwater navigation.
One notable example is the Loon Copter, a quadrotor-based UAAV capable of VTOL, as proposed by \cite{alzu2018loon}. The Loon Copter features a unique design that allows it to perform aerial flight, operate on the water surface, and dive underwater using a single set of motors and propellers. It is also equipped with a ballast system to control its buoyancy.  
Due to the low drag force in the air, these fixed-rotor UAAVs can effectively achieve horizontal movement by adjusting their attitude. However, the higher resistance makes it less effective to achieve horizontal movement through attitude changes in water, often requiring larger rotation angles to generate the desired movement.
\cite{9196687, 10160899, 10814088} introduced lightweight tilt-rotor UAAVs capable of VTOL from water surface. The thrust direction can be modified by tilting four rotors with a single servo or four servos. The thrust vectoring capability enables these vehicles to generate thrust both horizontally and downwards without the need to rotate their body.

UAAVs with delta-wing or monoplane configurations possess the long-range endurance of fixed wings. However, these vehicles have poor underwater capabilities or limited maneuverability on the water surface.
The SUWAVE, developed by \cite{7762760}, consists of a delta wing and a central body. When the central body is released by a latch and rotates to a vertical position due to gravity, the propeller on the central body generates thrust to lift the wing, achieving a passive vertical takeoff. To address the susceptibility of the passive strategy in \cite{7762760} to wave impacts, \cite{9120363} adopted an active strategy to alter the direction of the propeller, enabling autonomous takeoff. Although the SUWAVE series prototypes can take off from the water surface, they do not possess underwater maneuverability. Additionally, since the entire body remains on the water surface, their surface maneuverability is relatively constrained.

On the other hand, the CONOPS, a monoplane UAAV, is capable of repeated cross-medium operations and low-energy hovering, with a maximum surface speed of 1.2 m/s and a maximum underwater speed of 0.89 m/s\cite{8039429}. The Dipper, which features a new propulsion system with a maximum underwater speed of 3 m/s, is proposed by \cite{rockenbauer2021dipper}. The system allows for independent control of the underwater and aerial propellers by changing the motor direction.
The Longbow II adopts a novel foldable wing design, with a maximum surface speed of 1.05 m/s and a maximum underwater speed of 1.23 m/s\cite{sun2024design}. These monoplane UAAVs are capable of maneuvering underwater, but their mobility on the surface is still limited.

Hybrid UAAVs are designed to integrate various platform configurations to take advantage of these designs. For instance, a hybrid UAAV can integrate a multi-rotor configuration with an unmanned underwater vehicle(UUV), a fixed wing, an underwater glider, or a float\cite{jin2024nezha, 10577093, qin2023aerial, wang2024design, 10844691}. Additionally, it can combine a fixed-wing configuration with a glider, surface sailboat, or hydrofoil boat\cite{lyu2022toward, 8733846, zou2023design}. 

According to Table~\ref{tb:performance_paras}, it is clear that the existing prototypes lack the ability for rapid maneuverability both on the water surface and underwater. This limitation restricts their capacity to quickly cover large areas of water and transition between different bodies of water. This letter presents a new solution for water-air motion modes by designing a novel aquatic-aerial robot, SurfAAV, which possesses efficient maneuverability both underwater and on water surface, as well as in the air. The SurfAAV achieves a maximum surface speed of 7.96 m/s and a maximum underwater speed of 3.1 m/s. Moreover, experimental validation has demonstrated that the SurfAAV can successfully glide on water surface and take off from the water surface.

The remaining sections of this letter are organized as follows. Section II introduces the multi-modal locomotion concept of the SurfAAV. In Section III, the structural design and electrical system of the SurfAAV are elaborated. Section IV analyzes the simulation results of the prototype in computational fluid dynamics(CFD). The controllers used by the proposed prototype during the phases of submerging, gliding, and flying are discussed in Section V. Section VI presents the experiments conducted on underwater navigation, surface gliding, and the transition from water to air. Finally, Section VII concludes this paper.

\begin{figure*}
    \centering
    \includegraphics[width=0.9\linewidth]{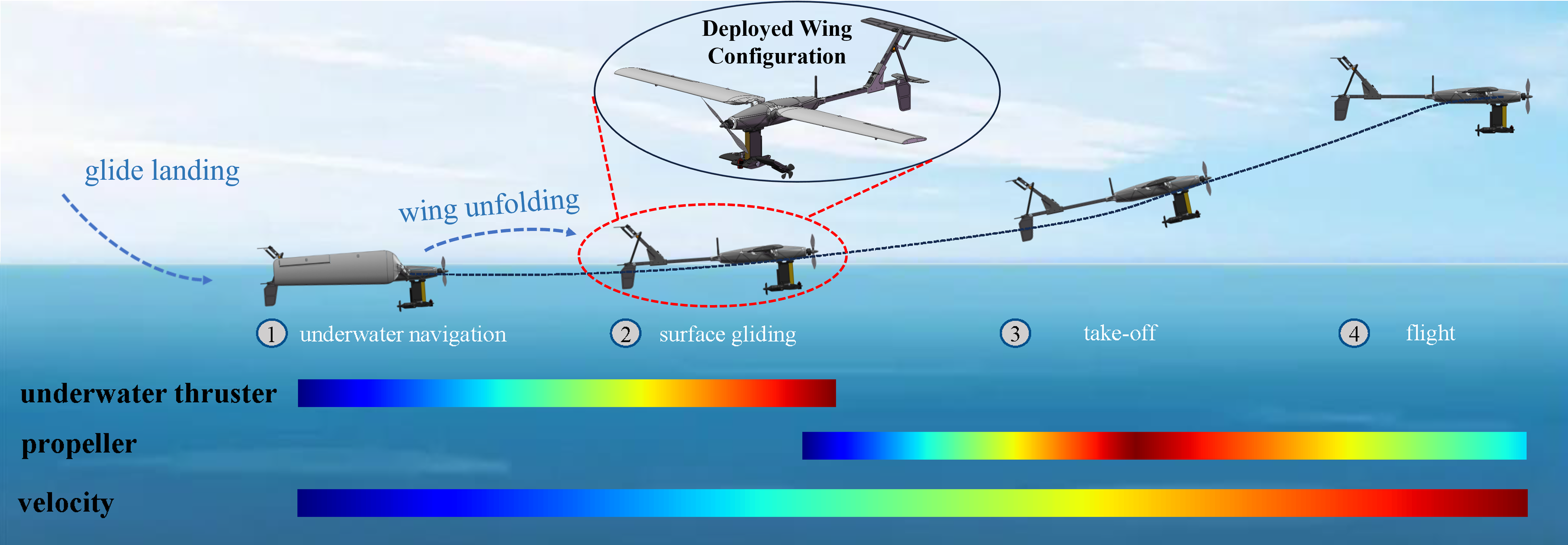}
    \caption{Schematic diagram of the multi-modal locomotion of the SurfAAV. The SurfAAV glides landing into water and transitions to the underwater navigation state to perform tasks. After completing these tasks, the vehicle returns to the water surface to take off and fly to another body of water. The surface takeoff is achieved by accelerating the underwater thruster and the aerial motor simultaneously.}
    \label{fig:multi-modal concept}
\end{figure*}

\begin{table}[h]
    \centering
    \label{tb:performance_paras}
    \caption{Performance parameters of present UAAV prototypes} 
    \begin{threeparttable}
        \begin{tabular}{c c c}
            \hline
            \textbf{Prototype} & \makecell{\textbf{Underwater speed} \\ \textbf{(m/s)}} &  \makecell{\textbf{Surface speed} \\ \textbf{(m/s)}} \\ \hline
            
            \makecell{\textbf{SurfAAV} \\ (our proposed)} & $\mathbf{3.1}^{\star}$ & $\mathbf{7.96}^{\star}$ \\ \hline
            
            \makecell{Loon Copter\textsuperscript{\cite{alzu2018loon}} \\ (2018)} &  0.5 &  1 \\ \hline

            \makecell{CONOPS\textsuperscript{\cite{8039429}} \\ (2018)} & $0.89^{\star}$ & $1.2^{\star}$ \\ \hline

            \makecell{Dipper\textsuperscript{\cite{rockenbauer2021dipper}} \\ (2021)} & 3 & \textbackslash \\ \hline

            \makecell{Nezha-B\textsuperscript{\cite{10577093}} \\ (2024)} & $1.67^{\star}$ & \textbackslash \\ \hline
 
            \makecell{SailMAV\textsuperscript{\cite{8733846}} \\ (2019)} & \textbackslash & $6^{\star}$ \\ \hline

            \makecell{QianXiang II\textsuperscript{\cite{zou2023design}} \\ (2023)} & \textbackslash & 4 \\ \hline
              
            \makecell{Diving Hawk\textsuperscript{\cite{wang2024design}} \\ (2024)} & $0.49^{\star}$ & \textbackslash  \\ \hline

            \makecell{WuKong\textsuperscript{\cite{10814088}} \\ (2025)} & \textbackslash & $2.15^{\star}$ \\ \hline
            
            \makecell{AquaAAV\textsuperscript{\cite{10844691}} \\ (2025)} & \textbackslash & 2.2\\ \hline
        \end{tabular}
        \begin{tablenotes}
          \footnotesize
          \item Note: The superscript $\star$ denotes the maximum speed.
        \end{tablenotes}
     \end{threeparttable}
\end{table}

\section{Multi-modal Locomotion Concept}
Current UAAVs exhibit limited surface and underwater maneuverability, as indicated in Table~\ref{tb:performance_paras}. The main reason for the reduced maneuverability of these UAAVs is that, when these robots operate on water surface, their entire body is submerged. This results in a large wetted surface area, which significantly increases drag on these robots. Additionally, an appropriate underwater propulsion system has not been adopted.

The SurfAAV combines the benefits of hydrofoil boats, unmanned underwater vehicles(UUVs), and fixed-wing aircraft, integrating the high-speed gliding mode as a hydrofoil boat on water surface, the underwater navigation mode as an UUV, and the flight mode as a fixed-wing aircraft. The multi-modal locomotion process of the SurfAAV is illustrated in Fig.~\ref{fig:multi-modal concept}. During the process, the SurfAAV transitions through five phases: underwater navigation, high-speed gliding on water surface, takeoff from the water surface, aerial flight, and glide landing. 
When submerged underwater, the vehicle is propelled by underwater thrusters and underwater ailerons. An autonomous depth control system allows the robot to reach the specified depth during underwater navigation. When it is necessary to move to another body of water to perform a task, the robot first emerges from underwater to the water surface by adjusting the underwater thrusters and the underwater ailerons, transitioning into gliding mode. Subsequently, it switches to flying mode by taking off from the gliding mode.
During the flight phase, a standard fixed-wing controller is used. The robot is powered by a 6S Li-Po battery with a capacity of 1800mAh. Finally, the vehicle returns to water surface to execute its next task. 

The transition of the SurfAAV to flight is composed of three main parts, as shown in Fig.~\ref{fig:multi-modal concept}. First, the underwater thrusters and the underwater ailerons are turned on, allowing the vehicle to glide slowly on the water surface. SurfAAV behaves as a hydrofoil at this stage, with its entire body above the water surface while the underwater thrusters and the underwater ailerons remain submerged, thus reducing the wetted surface area of the robot. Second, the aerial propeller is turned on as the gliding speed increases. The underwater thrusters and aerial propeller accelerate simultaneously. When the gliding speed reaches the minimum take-off speed, the SurfAAV transitions into flight mode, with a maximum flight speed of 22.87m/s. During the take-off process, the only part of the vehicle that contacts the water is the tail section. Finally, the underwater thrusters and underwater ailerons are turned off when the vehicle has completely left the water surface. In the flight stage, SurfAAV behaves as a fixed-wing aircraft.

\section{Mechanical and System Design}
In this section, the SurfAAV is presented. The structural design and electronic layout of this vehicle are also introduced.

\subsection{Structure Overview}
\begin{figure}
    \centering
    \includegraphics[width=0.9\linewidth]{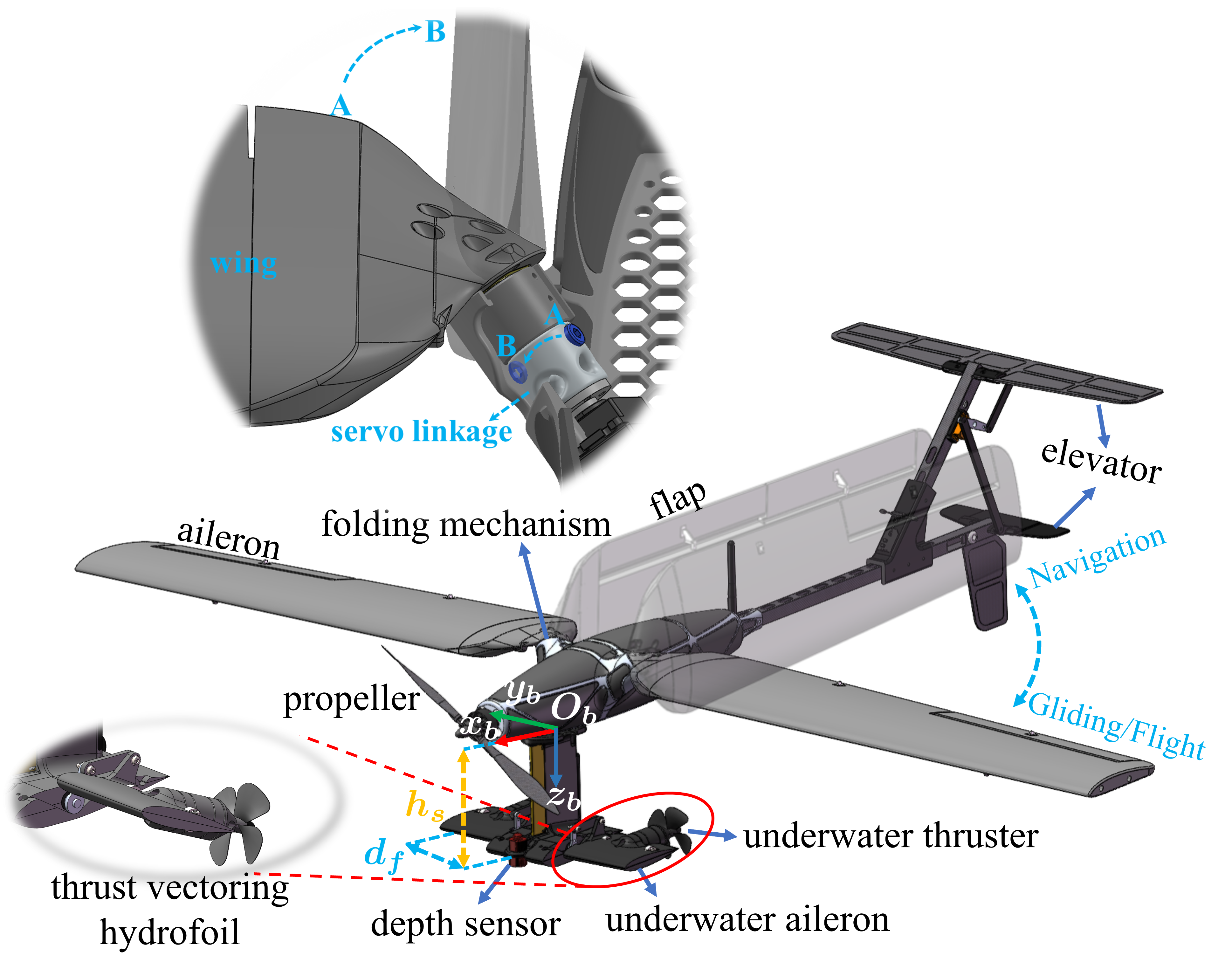}
    \caption{Structure diagram of the SurfAAV. The body frame is defined as $\mathcal{F}_b:\{O_b - x_b y_b z_b\}$. A. Wing configuration in flight and surface gliding modes, where the wing is in an unfolded state. B. Wing configuration in underwater navigation mode, where the wing is in a folded state.}.
    \label{fig:schematic_diagram}
\end{figure}

The SurfAAV, which is composed of an aerial section and an underwater section, is designed as illustrated in Fig.~\ref{fig:schematic_diagram}. The aerial section features a fixed-wing configuration, with the fuselage constructed from aluminum alloy as the framework and covered with skins. The wings are made from expanded polypropylene(EPP) materials wrapped around a wing spar, with carbon fiber tubes serving as the spar. The wings are attached to the folding mechanism using aluminum alloy to ensure structural rigidity and stability when the wings are folded. The dual-layer tail is designed with a flat-plate airfoil, made of hollow carbon fiber plates that are filled with lightweight polylactic acid(PLA) material. The connection between the fuselage and the tail wings is achieved using carbon fiber tubes. The selected aerial motor is the T-MOTOR V3120 KV700, which is paired with a 10-inch foldable propeller that has a pitch of 6 inches.

The underwater section primarily consists of underwater thrusters, underwater ailerons, a vertical strut, and servos. The vertical strut and underwater ailerons are 3D printed using PLA material, with the underwater ailerons functioning as all-moving control surfaces. These servos are embedded within the vertical strut, and an underwater thruster, model APISQUEEN X3, is mounted at the trailing edge of an underwater aileron, with a propeller size of 60mm. When the all-moving underwater aileron is actuated by the servo, the underwater thruster generates thrust vectoring.

\subsection{Wing Folding Mechanism}
Due to the exceptional multi-modal locomotion capabilities of flying fish, the fuselage shape of the SurfAAV has been designed based on their morphological characteristics. Most flying fish exhibit a spindle-shaped body with a distinct rounded-square cross-section, where their flattened belly contributes to additional lift generation\cite{deng2019numerical}. Inspired by this unique structure, the SurfAAV adopts a smooth spindle-shaped fuselage with a rounded-square cross-section. Furthermore, we designed a folding wing mechanism inspired by the physiological structure of flying fish pectoral fins, as shown in Fig.~\ref{fig:schematic_diagram}. The design features a simple and reliable structure that requires only a single servo to drive the folding of one wing, without the need for additional mechanical support. When folded, the wings avoid interference with the fuselage, enabling the use of larger wing areas to enhance flight efficiency and payload capacity. The key to SurfAAV’s wing-folding capability lies in the folding mechanism, which ensures that the wings are perpendicular to the fuselage's longitudinal axis when unfolded and parallel to it when folded.

\subsection{Thrust Vectoring Hydrofoil Design}
Traditional hydrofoil boats use hydrofoils to lift the hull out of the water, reducing the wetted surface area and thereby decreasing drag to achieve high-speed gliding\cite{suastika2022resistance}. However, the hydrofoils on these boats are usually fixed in direction and lack dynamic angle adjustments, limiting their effective working speed range. At low speeds, the hydrofoils cannot generate sufficient lift. During high-speed gliding, the lift produced by the hydrofoils is only enough to counteract gravity, enabling the hull to rise. 

In multi-domain missions, cross-domain robots equipped with traditional hydrofoils\cite{zou2023design} require a buoyancy adjustment system when transitioning to a submerged state. This paper proposes a novel differential thrust vectoring hydrofoil for the first time to enable cross-domain robots to achieve cruising, gliding, and takeoff capabilities simultaneously. The hydrofoil differs from traditional hydrofoils by using a fixed-wing profile as the shape of an underwater aileron. The underwater aileron is controlled by a servo that allows it to rotate. The lift and drag characteristics of the underwater aileron can be modified by adjusting the deflection angle of the underwater aileron. The optimal choice of wing profile is not unique and is influenced by various factors such as overall shape, camber, and thickness\cite{basinger2022design}. In this paper, we select the NACA 0016 profile as the wing shape for the underwater aileron. Subsequently, an underwater thruster is mounted on the underwater aileron, allowing the SurfAAV to rapidly lift its body during low-speed gliding. This is achieved by utilizing the thrust generated by the underwater thruster, which effectively reduces drag. This design also eliminates dependence on a buoyancy adjustment system to control the robot's ascent and descent.
When the SurfAAV needs to perform a submerging task, servos tilt underwater ailerons downward, enabling underwater thrusters to generate vertical downward thrust components, thereby achieving descent.

The height of the hydrofoil strut is 12.62 cm, which is not the same height as the horizontal tail. The height difference between the underwater ailerons and the horizontal tails allows the SurfAAV to raise its nose while gliding. The angle between the SurfAAV and water surface is about $12.4^{\degree}$, and this attitude caused by the angle is beneficial for the SurfAAV's takeoff from the water surface. The pitch angle of the SurfAAV during high-speed gliding on the water surface can be estimated to be approximately $12.4^{\degree}$ based on the vertical height of the hydrofoil and the horizontal distance from the horizontal tails to the strut. In practice, since the center of mass of the SurfAAV is located between the strut and the horizontal tails, and the submerged volumes of the underwater ailerons and the horizontal tail are different, the actual takeoff angle is slightly greater than $12.4^{\degree}$.

\subsection{Electrical System}
\begin{figure}
    \centering
    \includegraphics[width=0.9\linewidth]{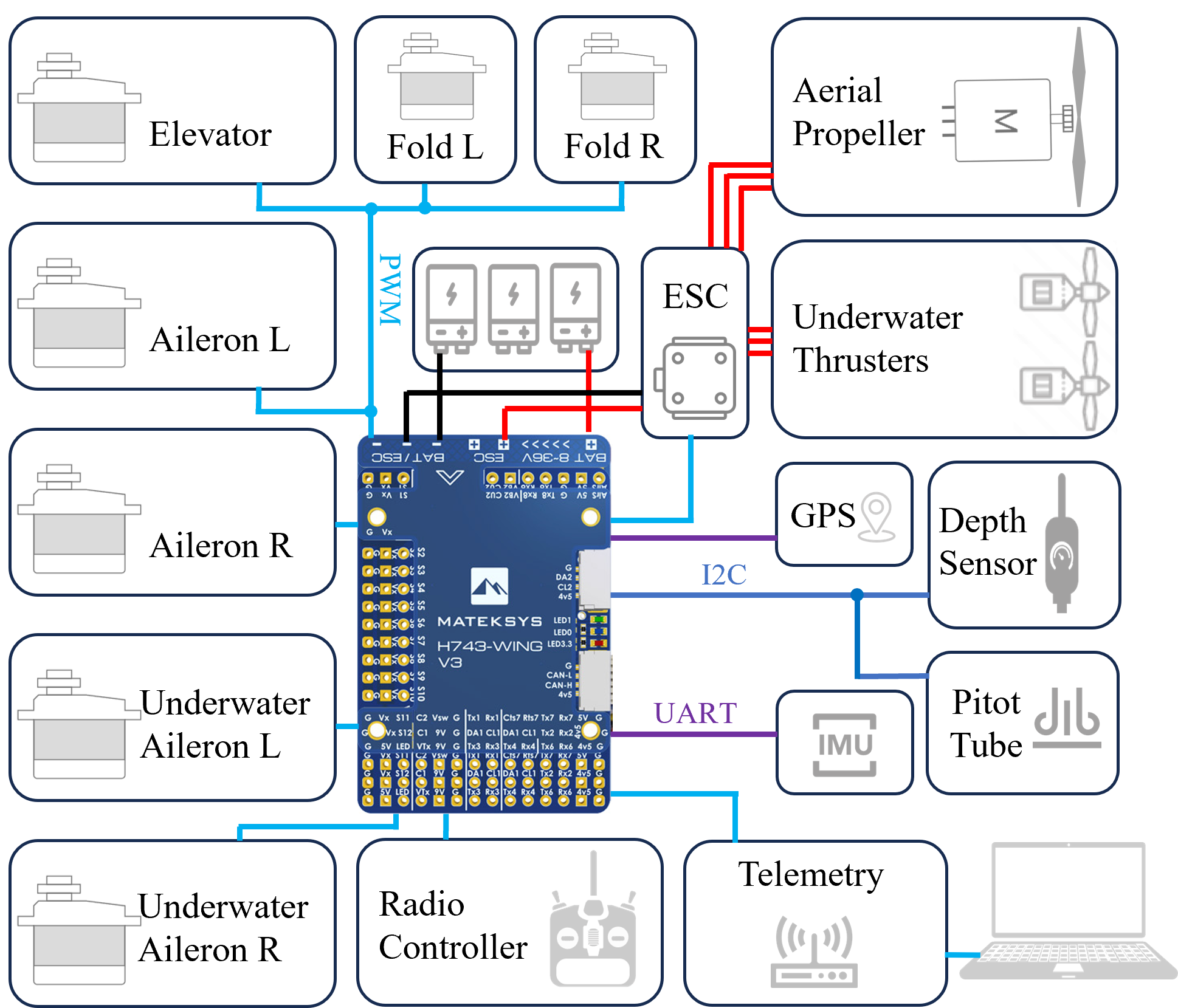}
    \caption{Avionics system layout of SurfAAV.}
    \label{fig:electronic_layout}
\end{figure}

The electrical structure of the SurfAAV is mainly divided into three modules: the propulsion module, the control module, and the energy module. As shown in the Fig.~\ref{fig:electronic_layout}, the propulsion module is further separated into aerial and underwater sections. The aerial section consists of an aerial thruster and five servos. Among them, two waterproof servos are used to control the flaps to increase lift during low-speed flight, with the initial angle of the flaps set to $15^{\degree}$. The ailerons are controlled by two servos, while the tails are operated by one servo. In the underwater section, there are two underwater thrusters and two waterproof servos that adjust the thrust direction of the thrusters. These thrusters are controlled by a four-in-one electronic speed controller(ESC), which adjusts their motor speed based on pulse-width modulation(PWM) control signals received from the control module.

The control module consists of an embedded flight control board and sensors. The Matek H743-wing V3 is chosen as the embedded flight control board, running the PX4 firmware. This open-source firmware allows for customizable functional programming to meet the demands of task switching\cite{7140074}. The board features an ARM Cortex-M7 core that operates at a maximum frequency of 480 MHz. It outputs control signals to eleven actuators and collects data from all connected sensors. A series of sensors are embedded in the control loop, including a global positioning system(GPS), a depth sensor, an inertial measurement unit(IMU) and a pitot tube, to provide the data support necessary for precise control. The selected GPS model is the CYCLONE M1018C, which provides the robot's position in the air and connects to the controller via the universal asynchronous receiver transmitter. Since the vehicle cannot receive satellite signals when in underwater navigation mode, a depth sensor, the MS5837, is used to measure underwater depth. All operational data is stored via a serial peripheral interface on an SD card. A 3DR radio transmission module X7, operating at a communication frequency of 915 MHz, is used to facilitate wireless data communication between the ground station and the flight controller, with a baud rate set at 115200 to ensure effective communication.

The power module mainly consists of a 6S LiPo battery that supplies power to the other modules. These sensors in the control module are primarily powered by a 5V voltage, which is obtained by converting 25.2V through a voltage regulation module. The output voltage of the 12-channel PWM signal is 5V. Waterproofing for the entire system is achieved using silicone sealant and adhesive heat shrink tubing. To ensure reliability and safety underwater, the flight control board and the four-in-one ESC are housed inside a waterproof sealing box.

\section{Aerodynamic Analysis}

\subsection{Simulation Setup}
A simulation analysis was conducted to examine the aerodynamic characteristics of the SurfAAV using STAR-CCM+ 18.02.008-R8, based on the Reynolds-averaged Navier–Stokes equations(RANS) and the K-Omega SST turbulence model. First, details that had minimal impact on fluid flow analysis were removed through simplified model processing and curvature continuity techniques to avoid mesh errors, such as negative volume errors, during the mesh generation process of the original model. Next, a rectangular fluid computational domain was established, and the corresponding boundary conditions were set. To ensure the accuracy and validity of the simulation results, a mesh independence verification was performed to determine the appropriate number of mesh divisions. A refined prism layer meshing technique was employed to maintain the y+ value in the boundary layer within six layers, ultimately optimizing the global mesh size to between 3.5 million and 4 million elements.

The aerodynamic simulation primarily investigated the flow field distribution characteristics under typical flight conditions, with an angle of attack(AoA) range from $-10^{\degree}$ to $20^{\degree}$ and a speed of 20 m/s. Through CFD simulation, lift and drag characteristic curves were obtained, and a quantitative analysis of the impact of the robot's shape design on aerodynamic efficiency was conducted. Additionally, the aerodynamic stability of the SurfAAV at different angles of attack was verified through the moment coefficient curve.

\subsection{Numerical Simulation Results}
\begin{figure*}
    \centering
    \includegraphics[width=0.85\linewidth]{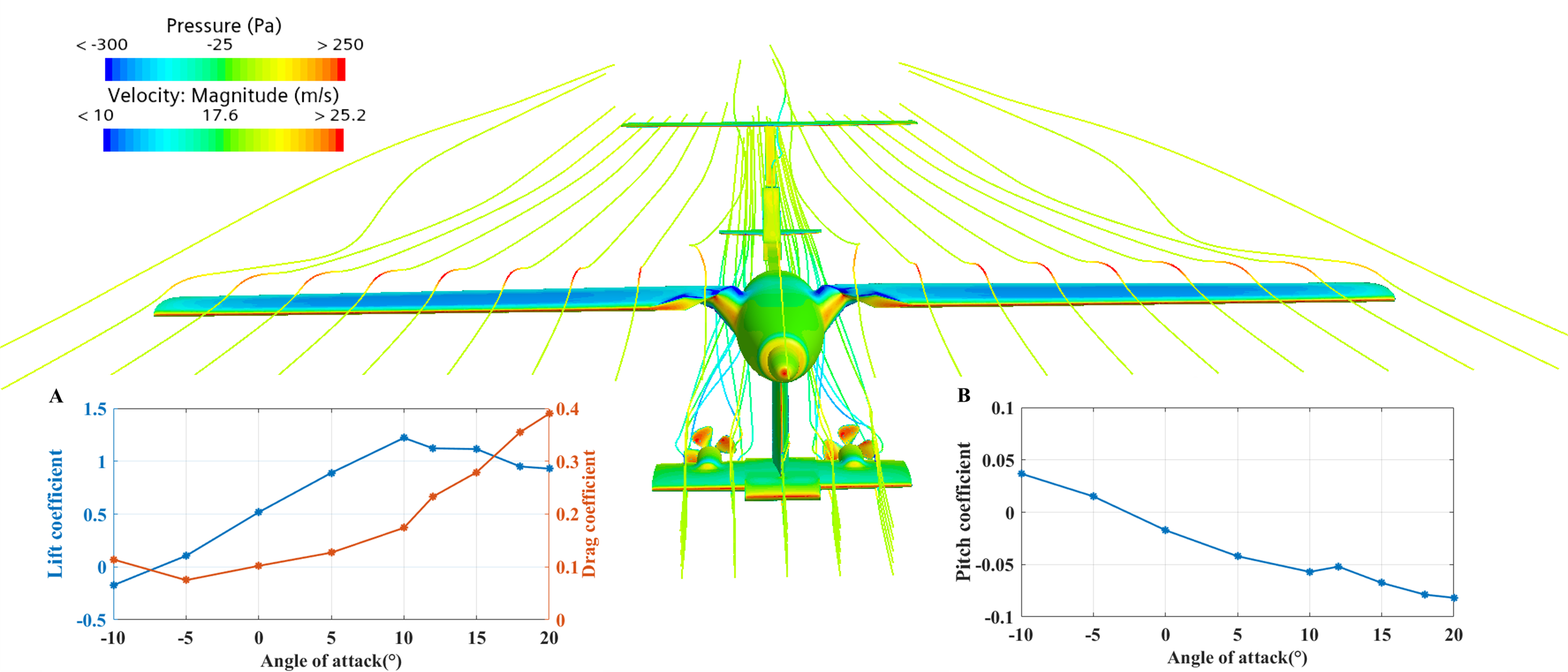}
    \caption{Pressure contour map of the SurfAAV surface. The streamlines are colored based on the magnitude of the velocity. A. The lift coefficient (black line) and drag coefficient (yellow line) vary with the angle of attack, with the stall angle at 10°. B. The pitching moment coefficient about the center of gravity for the SurfAAV. The negative slope of the pitching moment coefficient's rate of change with respect to the angle of attack indicates longitudinal stability, except immediately after stall.}
    \label{fig:pressure_contour}
\end{figure*}

Fig.~\ref{fig:pressure_contour}A shows the curves of the lift coefficient ($C_L$) and drag coefficient ($C_D$) as a function of the AoA. When the AoA is negative, the lift coefficient is also negative, indicating that the wing is stalled and cannot generate effective lift. As the AoA increases from $0^{\degree}$ to $10^{\degree}$, lift increases significantly, while drag also rises. During this range, the performance of the SurfAAV is relatively ideal and remains within a safe range. At an AoA of 10°, the lift coefficient reaches its maximum value of 1.221, with a drag coefficient of 0.174, resulting in a maximum lift-to-drag ratio of 7.02. At this angle, the aerodynamic efficiency of the robot is optimal, meeting the energy consumption and lift balance requirements for cruise flight. However, as the AoA exceeds $10^{\degree}$, airflow begins to separate. Although lift decreases, drag increases significantly, causing the SurfAAV to stall and exhibit unstable characteristics.

Fig.~\ref{fig:pressure_contour}B illustrates the pitch moment coefficient around the center of mass as a function of the AoA. According to conventional static stability analysis, when the SurfAAV is in a statically stable state, the pitch moment coefficient generally shows a trend of increasing negative values as the AoA increases, producing a nose-down moment to counteract further increases in the AoA. However, as the SurfAAV approaches or exceeds the stall angle, the lift coefficient can drop sharply due to flow separation, which may cause a sudden change in the pitch moment coefficient. 
In the small AoA phase range from $0^{\degree}$ to $10^{\degree}$, the pitching moment coefficient($C_m$) decreases linearly as the AoA increases. During this period, the airflow remains well attached, the horizontal tail is effective, and the SurfAAV is statically stable. As the AoA exceeds the stall angle of $10^{\degree}$, the rate of change of $C_m$ becomes positive, and an inflection point begins to appear. In the post-stall phase, severe flow separation occurs, leading to a decrease in the efficiency of the horizontal tail, which may result in the SurfAAV becoming unstable.

\section{Multimodal Motion Control}
\begin{figure}
    \centering
    \includegraphics[width=0.9\linewidth]{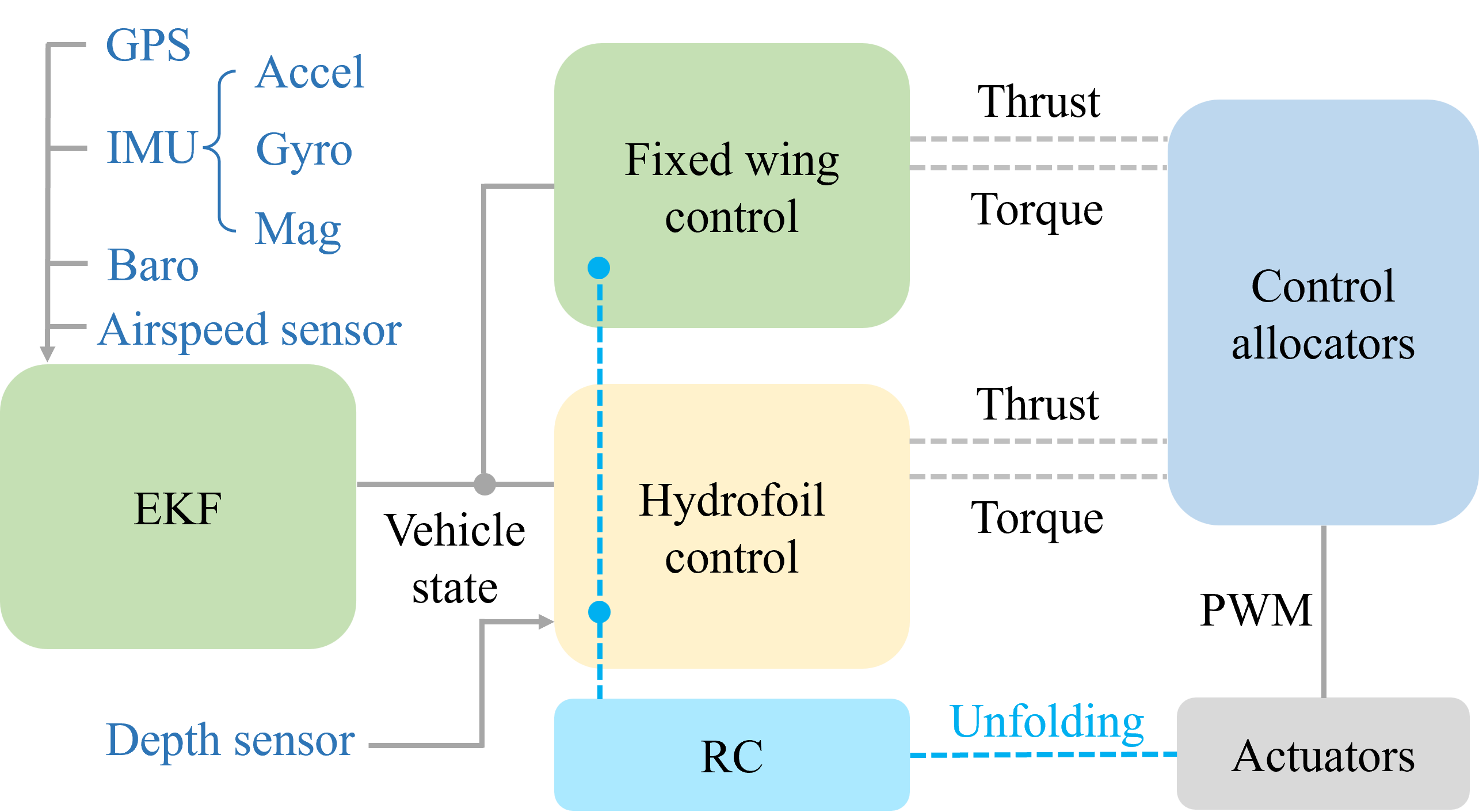}
    \caption{Control Block Diagram of the SurfAAV.}
    \label{fig:control_diagram}
\end{figure}

The SurfAAV requires different control systems for various modes, including flight mode, gliding mode, and submerging mode. Fig.~\ref{fig:control_diagram} illustrates the core control components necessary for our prototype. In flight mode, the SurfAAV operates like a fixed-wing aircraft, so standard fixed-wing controls can be applied, meaning that the ailerons, elevators, and propellers are controlled accordingly. The fixed-wing controller can be available from the open-source flight control system PX4.

Once the SurfAAV lands on the water, the hydrofoil controller is enabled through a radio signal. This custom control module follows the specifications and standards of the PX4 modules and is primarily divided into attitude control and depth control. The attitude control employs control laws similar to those used in fixed-wing aircraft and listens to the uORB topic \emph{vehicle attitude} provided by the EKF module. The depth control is achieved via the vertical force component generated by the hydrofoil, with the control law as follows
\begin{equation}
    f_{zd} = k_p e_h + k_i \int e_h dt + f_{g}
\end{equation}
where $e_h = h_d - h$, the depth h is the data obtained from the depth sensor, and $f_{g}$ is the gravity compensation term.

Using the hydrofoil controller, the desired forces and moments $\begin{bmatrix} f_{xd}, f_{zd}, \tau_{xd}, \tau_{yd}, \tau_{zd} \end{bmatrix}^T$ can be obtained, and the desired wrench can be allocated to the virtual force components of the hydrofoil using the following equation
\begin{equation}
    \begin{aligned}
    \begin{bmatrix}
        f_{xd} \\
        f_{zd} \\
        \tau_{xd} \\
        \tau_{yd} \\
        \tau_{zd}
    \end{bmatrix} = 
    \begin{bmatrix}
        1   &  0  &  1  &   0  & 0 \\
        0   &  1  &  0  &   1  & 0 \\
        0   & d_f &  0  & -d_f & 0 \\
        h_s &  0  & h_s &   0  & 1 \\
       -d_f &  0  & d_f &   0  & 0
    \end{bmatrix}
    \begin{bmatrix}
        F_{xbr} \\
        F_{zbr} \\
        F_{xbl} \\
        F_{zbl} \\
        m_e
    \end{bmatrix}
    \end{aligned}
    \label{eq:allocation_mat}
\end{equation}
where $d_f$ and $h_s$ are defined in Fig.~\ref{fig:schematic_diagram}.

Considering only the thrust $F_T$ from the underwater thruster and the hydrodynamic forces ($F_L, F_D$) acting on the underwater aileron, the forces acting on the underwater aileron can be decomposed orthogonally in the body coordinate system $\mathcal{F}_b$, resulting in
\begin{equation}
    \begin{aligned}
        F_{xbi} &= F_{Ti} c_{\delta_i} + F_{Li} s_{\alpha} - F_{Di} c_{\alpha} \\
        F_{zbi} &= F_{Ti} s_{\delta_i} - F_{Li} c_{\alpha} - F_{Di} s_{\alpha} \\
    \end{aligned}
    \label{eq:orth_decomp}
\end{equation}
where $F_{*i} (i = r, l)$ represents the force acting on the right and left underwater ailerons, respectively. $\alpha$ denotes the angle of attack(AoA). In Eq.~\eqref{eq:orth_decomp}, there are only two unknown variables, namely thrust $F_{Ti}$ and servo deflection angle $\delta_i$, which can be solved using the Newton method.

\section{Experimental Results}

\subsection{Underwater Navigation}
\begin{figure}
    \centering
    \includegraphics[width=0.8\linewidth]{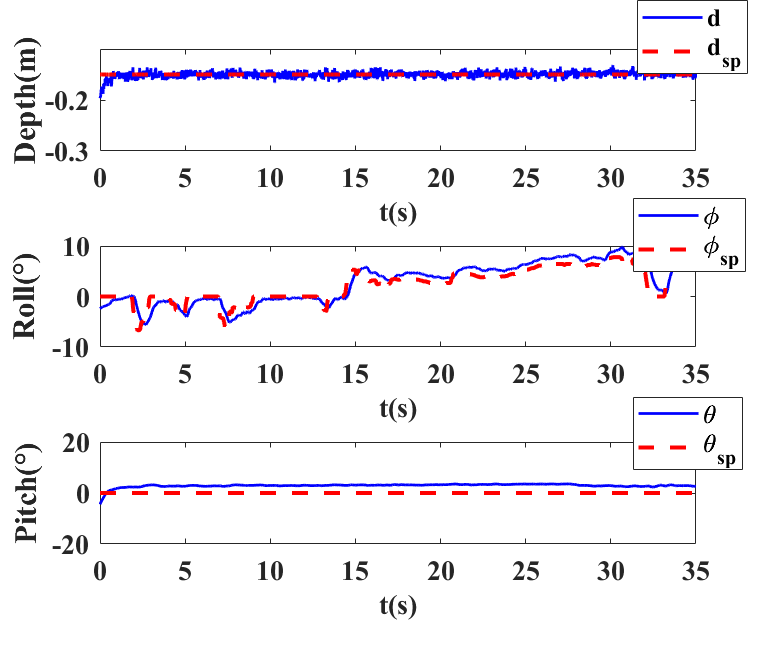}
    \caption{Depth-maintained underwater cruising. During underwater cruising, the wings of the SurfAAV are folded, and real-time depth measurement data is provided by the depth sensor.}
    \label{fig:underwater_navigation}
\end{figure}

During the underwater navigation process, the wings of the SurfAAV are in a folded state, while the depth sensor provides real-time depth measurement data. The experimental results for the depth-hold cruising of the SurfAAV are presented in
Fig.~\ref{fig:underwater_navigation}. The target depth is set at 0.15 meters underwater. It can be observed that the SurfAAV can achieve autonomous depth control by using the differential thrust vectoring hydrofoil from the results. The desired roll angle and pitch angle are provided by the RC. In this experiment, the effectiveness of the proposed hydrofoil control algorithm for autonomous depth maintenance was verified. The depth tracking error is approximately ±2 centimeters. There is a steady-state error of about 4 degrees in the pitch control loop, which may be due to the negative pitching moment generated by the lower tail wing not fully compensating for the positive pitching moment provided by the underwater thruster.

\subsection{Surface Gliding}
\begin{figure*}
    \centering
    \includegraphics[width=0.9\linewidth]{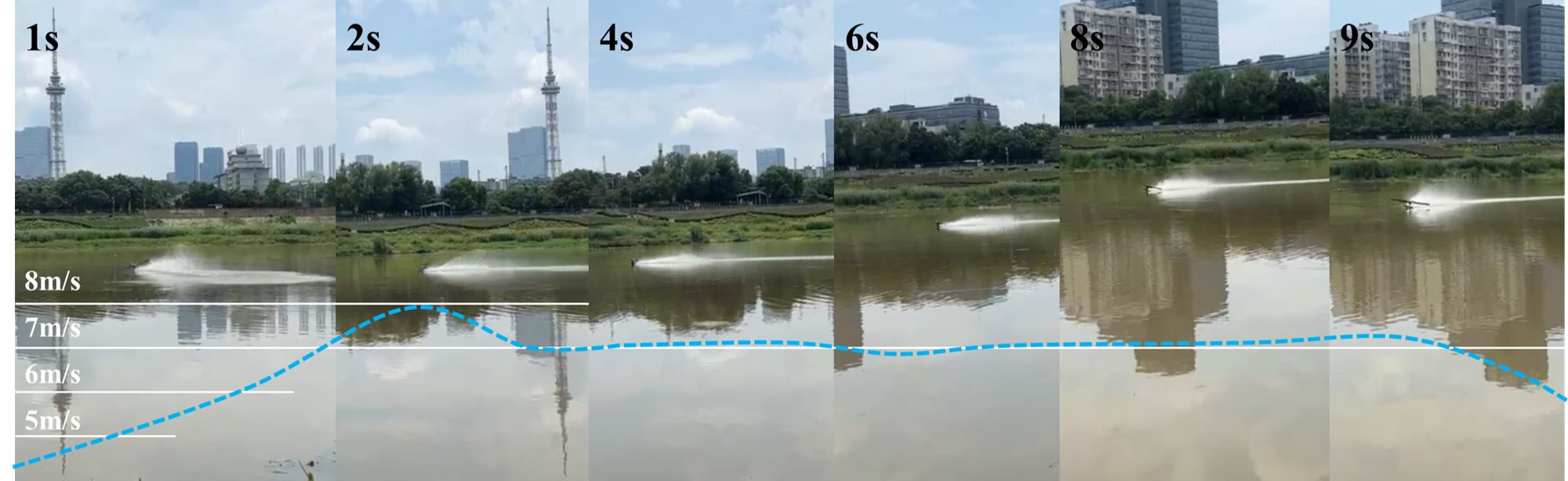}
    \caption{Keyframe screenshots of the surface gliding experiment on the Liuyang River and the gliding speed curve of the SurfAAV. The maximum surface gliding speed is 7.96 m/s.}
    \label{fig:surfing}
\end{figure*}

In the surface gliding experiment conducted on the Liuyanghe River in Changsha, we captured keyframe screenshots under calm water conditions, as illustrated in the Fig.~\ref{fig:surfing}. The experimental results showed that the SurfAAV reached a maximum speed of 7.96 m/s while gliding on the water surface. At low speeds, the lift generated by the underwater ailerons is insufficient to support the body of the SurfAAV above the water surface. However, the thrust from the underwater propellers can generate a vertical upward thrust component, causing the body of the SurfAAV to gradually rise out of the water surface. As the gliding speed increases, the wing surfaces of the underwater ailerons begin to generate additional lift, further elevating the body. Eventually, as the speed continues to rise, the entire hydrofoil emerges from the water. Simultaneously, due to the high-speed rotation of the underwater propellers near the water surface, the fluid velocity around the propellers gradually increases, resulting in the fluid pressure in that area dropping below atmospheric pressure, which draws in air and produces a quasi-cavitation effect\cite{Luo2020Thrust}.

\begin{figure}
    \centering
    \includegraphics[width=0.9\linewidth]{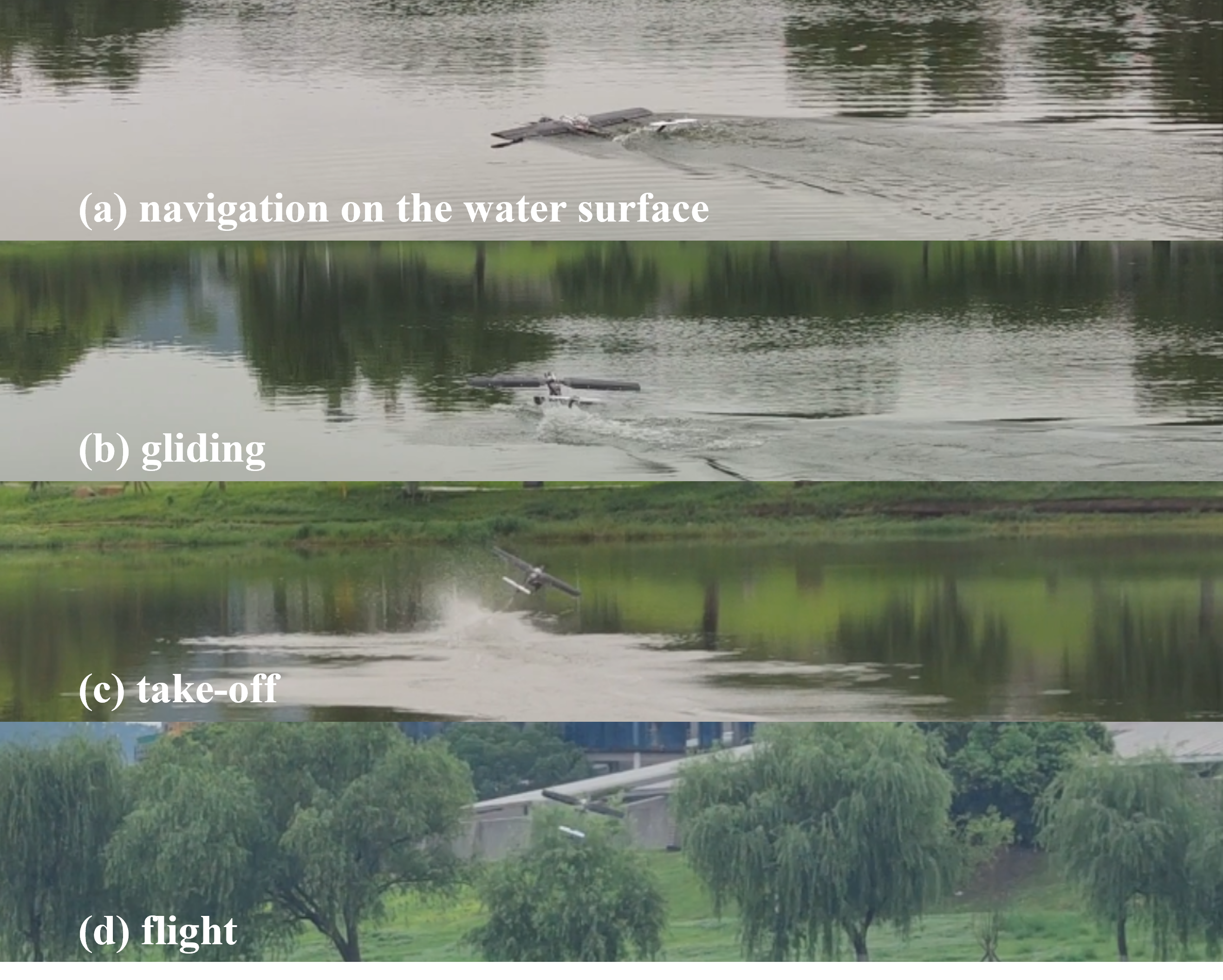}
    \caption{Transition from water to air.}
    \label{fig:water_to_air}
\end{figure}

\subsection{Transition From Water to Air}
An outdoor experiment was conducted to evaluate the transition of the SurfAAV from water to air, as illustrated in the Fig.~\ref{fig:water_to_air}, which shows the different stages of this process. First, the SurfAAV cruises on the water surface with its wings fully unfolded. Next, the throttle of the underwater propellers is increased, allowing the underwater ailerons to lift the body of the SurfAAV out of the water during high-speed gliding, thereby reducing the wetted area of the SurfAAV and minimizing drag. It can also be shown that the nose-up attitude of the SurfAAV, resulting from the height difference between the underwater ailerons and the tails, is beneficial for takeoff through this experiment. Once the body of the SurfAAV exits the water, the aerial propeller is activated, and the throttle is further increased for both the underwater propellers and the aerial propeller to enhance the gliding speed of the SurfAAV. As the speed exceeds the takeoff threshold, the entire SurfAAV is lifted from the water surface. Finally, after the underwater propellers are completely above the water surface, the throttle signal for the underwater propellers is turned off, and the SurfAAV transitions into aerial flight mode.

\section{Conclusions}
In this paper, we propose an innovative prototype design concept for a hybrid aquatic-aerial robot, which is capable of underwater cruising, surface gliding, and aerial flight. Thanks to the proposed novel differential thrust vectoring hydrofoil, SurfAAV can efficiently perform measurements both on the water surface and underwater, while utilizing its flight capability to move between different water bodies. The prototype of this robot has been successfully designed and implemented. Experimental validation demonstrates that the SurfAAV possesses the ability to cruise underwater, glide on the water surface, and fly in the air, successfully achieving take-off through gliding. Additionally, CFD results indicate that the SurfAAV is capable of stable flight. Its maximum gliding speed is 7.96 m/s, maximum cruising speed is 3.1 m/s, and maximum flying speed is 22.87 m/s. Furthermore, the hydrofoil controller has been found to be effective in maintaining a specific depth. The proposed robot is suitable for various application scenarios, such as three-dimensional monitoring of aquatic environments. Compared to multi-robot solutions, SurfAAV can be rapidly deployed, significantly improving the efficiency of water monitoring.

Future work will concentrate on the research of a hybrid control framework, further expanding it to achieve fully autonomous control. Additionally, we plan to implement a glide landing method for the SurfAAV's transition from air to water, which can effectively reduce the impact of water entry on the robot and minimize equipment damage. Finally, existing speed sensors struggle to balance range, size, and weight, making it challenging to provide suitable underwater speed measurement solutions for the small robot SurfAAV. To address this, we intend to develop a pressure-based underwater speed sensor suitable for small robots.





\ifCLASSOPTIONcaptionsoff
  \newpage
\fi

\bibliographystyle{IEEEtran}
\bibliography{ref/ref.bib}

%








\end{document}